\documentclass[subscriptcorrection,upint,varvw,barcolor=Goldenrod3,mathalfa=cal=euler,balance,hyphenate,french,pdf-a,nolists]{asmejour} %

\hypersetup{%
	pdfauthor={Yash Patawari Jain},                       		   	
	pdftitle={JMD - Design By Data - Material Dataset},                  	
	pdfkeywords={},
	pdfsubject = {},			
	pdflicenseurl={https://ctan.org/pkg/asmejour},
}

\JourName{Mechanical Design}

\begin{document}


\SetAuthorBlock{Yash \ Patawari Jain}{%
Department of Mechanical Engineering,\\ Carnegie Mellon University,\\
Pittsburgh, PA\\
ypatawar@andrew.cmu.edu
}

\SetAuthorBlock{Daniele Grandi}{Autodesk Research,\\
   San Francisco, CA\\
   daniele.grandi@autodesk.com} 

\SetAuthorBlock{Allin Groom}{%
   Autodesk Research,\\
   London, UK \\
    allin.groom@autodesk.com
}

\SetAuthorBlock{Brandon Cramer}{%
   Autodesk Research,\\
   Boston, MA \\
    brandon.stewart.cramer@autodesk.com
}

\SetAuthorBlock{Christopher McComb\CorrespondingAuthor}{Department of Mechanical Engineering,\\
   Carnegie Mellon University,\\
   Pittsburgh, PA\\
   ccm@cmu.edu} 


\title{MSEval: A Dataset for Material Selection in Conceptual Design to Evaluate Algorithmic Models}


\begin{abstract}
Material selection plays a pivotal role in many industries, from manufacturing to construction. Material selection is usually carried out after several cycles of conceptual design, during which designers iteratively refine the design solution and the intended manufacturing approach. In design research, material selection is typically treated as an optimization problem with a single correct answer. Moreover, it is also often restricted to specific types of objects or design functions, which can make the selection process computationally expensive and time-consuming. In this paper, we introduce \texttt{MSEval}, a novel dataset which  is comprised of expert material evaluations across a variety of design briefs and criteria. This data is designed to serve as a benchmark  to facilitate the evaluation and modification of machine learning models in the context of material selection for conceptual design.


\end{abstract}

\date{Version \versionno, \today}

\maketitle 


\begin{figure*}[h]
    \centering
    \includegraphics[width=\textwidth]{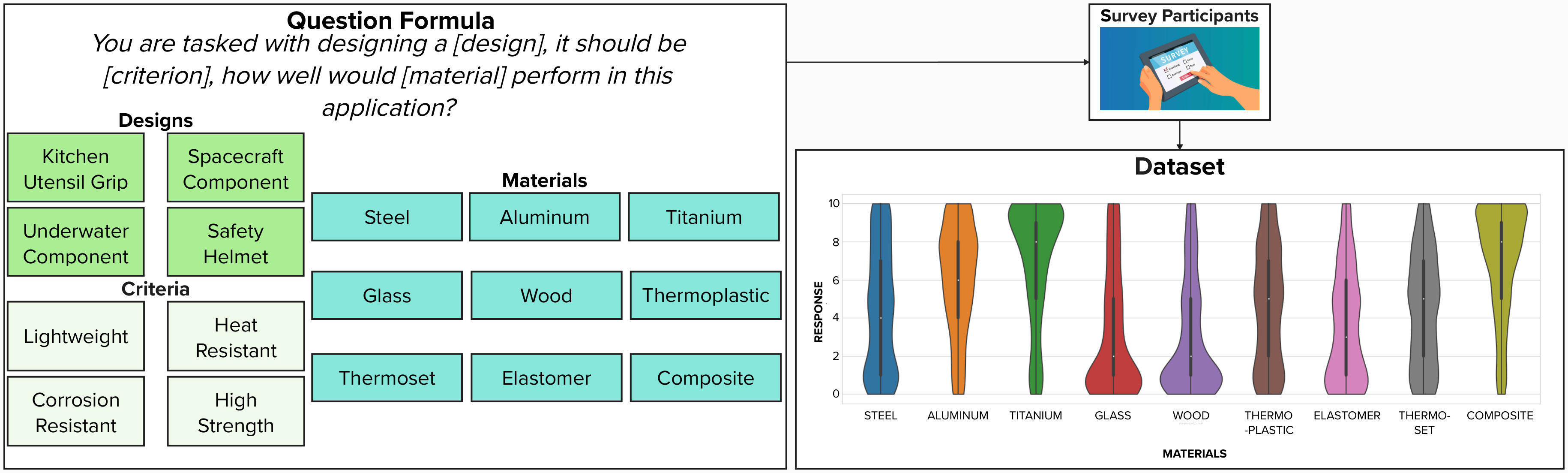}
    \caption{Overview of the method used to create the corpus of questions asked to survey participants. Completing the tasks requires fundamental abilities such as reasoning, critical thinking and knowledge of the material selection domain. Questions are ambiguous by design to invoke inherent requirements. The question formula is built through the factorial combination of four designs, four criteria, and nine materials. The violin plot shows the distribution of survey responses broken down by material.}
    \label{fig:overview}
\end{figure*}

\section{Introduction}
\label{sec:intro}
The process of material selection is critical across a wide variety of industries, ranging from manufacturing to construction. It involves making informed decisions about the materials used in various products, structures, and systems. Traditionally, material selection occurs after iterative design cycles, where design modifications and manufacturing methods are refined. However, this approach often neglects the crucial role that materials may play in shaping the performance, cost, and sustainability of the final product.

Historically, material selection has often followed a \textit{manufacturing first} approach in which designers and engineers tend to prioritize manufacturing feasibility over material suitability. Consequently, this bias can lead to suboptimal material choices, affecting product performance, durability, and overall quality.
Not all designers have extensive material expertise. Many lack the knowledge necessary to evaluate material properties, compatibility, and trade-offs effectively. Bridging this knowledge gap is essential for informed decision making.
Material selection decisions made early in the design process significantly influence downstream activities, so it is crucial to ensure that these decisions align with the project goals and constraints.

Typically, during the early design stage, material requirements are often expressed in the form of text and natural language. Designers describe desired properties, constraints, and performance criteria in qualitative terms, which then need to be interpreted and translated into specific material choices. This process can be challenging due to the nuanced and context-dependent nature of natural language descriptions.

The recent development of large language models (LLMs) in the machine learning domain \cite{brown_language_2020, vaswani_attention_2023, wei2022emergent, openai2023gpt4, touvron2023llama, driess2023palm} presents a novel opportunity to streamline the material selection process in the mechanical design domain. LLMs are a type of model capable of generating text. They have been used successfully in various domains, including design \cite{ma2023conceptual, zhu2023generative, jiang2022patent, picard2023concept, picard2024untrained, meltzer2024s, song2024multi}, where they have demonstrated the ability to generate innovative and creative visual and textual design solutions. As these AI systems continue to become more integrated into our daily lives, it is essential to effectively identify potential shortcomings, and ensure that they can handle complex, human-centric tasks effectively. Conventional evaluation benchmarks for LLMs frequently fall short in accurately assessing their overall capabilities for handling human-level tasks. As a result, there is an increasing demand for a human-centric benchmark that enables robust evaluation of foundational models within the context of tasks relevant to human reasoning and problem-solving \cite{zhong_agieval_2023}.


In this paper, we introduce an evaluation dataset curated from human survey responses that is designed to evaluate the abilities of LLMs for the task of material selection in conceptual design. This benchmark is curated from responses from a survey sent out to professionals working in the fields of material science, material selection, and design engineering, among others. The responses come from people working with varying levels of experience, and therefore capture the variance in the thought process and inherent considerations that come with different design tasks. By focusing on these different design tasks, our benchmark enables a more meaningful and contextual evaluation of the performance of the large language model in scenarios directly relevant to material selection. The overall approach is shown in figure \ref{fig:overview}.

The remainder of this paper is organized as follows. Section~\ref{sec:bg} reviews literature that is relevant to the dataset. Next, Section~\ref{sec:meth} reviews the the methodology used to generate the dataset as well as the organization of the dataset itself. Section~\ref{sec:pa}  discusses several possible ways in which this dataset might be used to support research. Finally, Section~\ref{sec:conc} summarizes the contributions and discusses limitations and future work. Ultimately, we aim to drive innovation in developing more reliable AI assistants that advance toward this niche domain by identifying areas for improvement and understanding the behavior of humans better. Our research underscores the importance of evaluating foundation models in the context of human-level material selection and provides a benchmark for such evaluations. We hope that our findings inspire further innovation and progress in the development and evaluation of large models in a niche task, ultimately leading to more reliable systems.


\section{Background}
\label{sec:bg}
In this section we review works related to material selection and its challenges and motivation to work on this problem.

\subsection{Significance and Challenges in Material Selection}
Material selection is a critical part of designing and producing any physical object. Material selection occurs in the early stages of the design workflow and maintains relevance beyond the useful life of a product. Materials directly influence the functionality, aesthetics, economic viability, manufacturing feasibility, and ultimately its environmental impact of a design \cite{zarandi_material_2011,bi2017energy,noauthor_materials_nodate}. 
M. F. Ashby is often cited for presenting a systematic approach to material selection through the use of bubble plots, known as "Ashby" diagrams, which allow a designer to evaluate up to two material properties to identify those materials that perform above a desired threshold \cite{ashby_selection_2004}. This approach requires an intimate understanding of a product's design intent, the design priorities (such as low mass), constraints (manufacturing process), and other requirements relevant to the object being designed (industry regulations).
In recent years, additional factors have become increasingly important to consider too. Sustainability, for example, is a growing global concern and manufacturing alone is reported to contribute significantly to resource consumption and greenhouse gas emissions. Thus, selecting materials with lower environmental impact, such as recycled content or those that require less energy to produce, aligns with ethical practices and growing consumer expectations \cite{kishita_checklist-based_2010,banu_joint_2024,ramani_integrated_2010}. Material availability is also becoming a critical consideration due to supply chain disruptions, geopolitical challenges, or regulations on material use. The growing complexity of design requirements does not reduce the implications of improper material selection which can lead to increased overall costs, product failure, or greater environmental harm \cite{albinana_framework_2012}. 

In product design, material selection is often decomposed into a general five-step procedure: (1) establishing design requirements, (2) screening materials, (3) ranking materials, (4) researching material candidates, and (5) applying constraints to the selection process \cite{bhat_aerospace_2018}. Charts of performance indices and material properties, called Ashby diagrams, are often used to visualize, filter, and cluster materials \cite{bhat_aerospace_2018,doi:10.1179/mst.1989.5.6.517}. 

Traditionally, material selection has relied heavily on engineering intuition and familiarity with existing materials. Particularly in industries with less prescriptive standards or specifications \cite{prabhu_favoring_2021,karandikar_approach_1992}.  Even with Ashby's systematic approach to material selection, the process is nontrivial and can still leave designers with uncertainty as to how well a candidate material will perform in reality \cite{ashby_materials_2011,ashby_selection_2004,doi:10.1179/mst.1989.5.6.517}. Data and knowledge are essential, without which a limited exploration of alternative or innovative options can occur, leading to suboptimal designs \cite{shiau_optimal_2009,hazelrigg_irrationality_1997}. Although established methods like Ashby's can guide designers and encourage them to consider a wider range of possibilities, material databases \cite{ullah_investigation_2020} cannot often account for the ever-growing universe of materials and broadening design considerations outlined thus far.
Selection can also be subjective, potentially overlooking promising new materials simply because designers are unfamiliar with them \cite{singh_subjective_2022,leontiev_how_2022}. Uncertainty regarding the performance of novel materials can further hinder their adoption. Furthermore, manufacturing innovations such as additive manufacturing are allowing previously unfeasible materials to now become viable options \cite{jelinek_design_2015}. This highlights the need for a data-driven approach to material selection, one that can objectively evaluate a broader range of options while considering the complex interplay of design requirements and provide information when selecting a particular option \cite{dong_survey_2017}. Machine learning and algorithmic models offer powerful tools for material selection \cite{yash2024material, yash2024material2}. By learning from vast datasets of past design experiences and material properties, models can provide valuable insights that would otherwise require extensive research or experimentation.

\subsection{Existing Evaluation Datasets}
To establish robust evaluation standards and effectively monitor model performance, reliable benchmarks play a pivotal role. While several well-known benchmarks exist for single-task and generalistic evaluation, they predominantly focus on assessing specific machine skills using artificially curated datasets. For instance, the SQuAD dataset \cite{rajpurkar2016squad} evaluates answer extraction ability, and the SNLI dataset \cite{bowman_large_2015} assesses natural language inference capability. Additionally, the GLUE \cite{wang_glue_2018} and SuperGLUE \cite{wang_superglue_2020} datasets serve as litmus tests for language models across various NLP tasks. However, these benchmarks often lack real-world applicability and fail to address complex reasoning abilities that align with human behaviors.

To bridge this gap, researchers have introduced novel datasets such as the MMLU \cite{hendrycks_measuring_2021} and AGI-Eval \cite{zhong_agieval_2023}. These datasets take a more holistic approach by collecting diverse subject data, guiding evaluation toward a human-centric perspective. By incorporating real-world scenarios and nuanced reasoning challenges, they provide a more comprehensive assessment of language models’ capabilities.

Although these benchmarks are comprehensive, they do not cover many niche applications that have unique requirements and challenges. Many niche benchmarks exist for various domains. BioASQ \cite{krithara_bioasq-qa_2023} is a well-known dataset for biomedical semantic indexing and question answering. It contains a large number of biomedical articles and associated questions designed to test models' ability to retrieve relevant biomedical information and provide precise answers. MedQA \cite{jin_what_2020} is a data set consisting of medical exam questions aimed at evaluating medical knowledge models. CaseLaw \cite{petrova_chinmusiqueoutcome-prediction_2022} dataset includes a collection of court cases and associated annotations for legal reasoning and case outcome prediction. The European Court of Human Rights (ECtHR) dataset \cite{ilias_chalkidis_ecthr-naacl2021_nodate} contains cases and rulings to evaluate models on the prediction of legal judgements and rationale generation. The FiQA dataset \cite{thakur2021beir} consists of financial question-answer pairs and sentiment analysis data related to financial markets. The FinCausal dataset \cite{noauthor_fincausal_nodate} is designed to evaluate causality detection in financial documents, which is crucial to understanding financial events and their impacts. SciQ \cite{SciQ} dataset includes multiple-choice questions derived from science textbooks, aimed at testing scientific knowledge. PubMedQA \cite{jin2019pubmedqa} contains questions and answers derived from PubMed abstracts, focusing on biomedical research queries. 

There exist multiple material properties databases such as the NIST Materials Data Repository \cite{noauthor_materials_nodate-2} and the Materials Project database \cite{noauthor_materials_nodate-1}, but they contain material properties and not how a human would interact in complex real-world design scenarios. Our goal is to create a dataset that could be used to bridge the gap between the thinking of a human designer/material selector and a machine learning model.

\section{Methodology}
\label{sec:meth}
In this section, we elaborate on the methodology that we used to collect data and the principles to make it accessible. In the first step \ref{sec:meth - online survey}, we collected materials selection perspectives from professionals through an online survey. Following the responses, we clean the responses to make two different variants of the dataset for different uses. In the final stage \ref{sec:meth - FAIR}, we elaborate on the FAIR principles.  

\subsection{Survey Design}
\label{sec:meth - online survey}
To collect data to act as the evaluation data set, we conducted an online survey among professionals in design, material science,  engineering, and related fields. Specifically, our goal was to get responses from people with varied experience in the field of material selection for mechanical design. 

\begin{figure}[htbp]
\centering

\includegraphics[width=\linewidth]{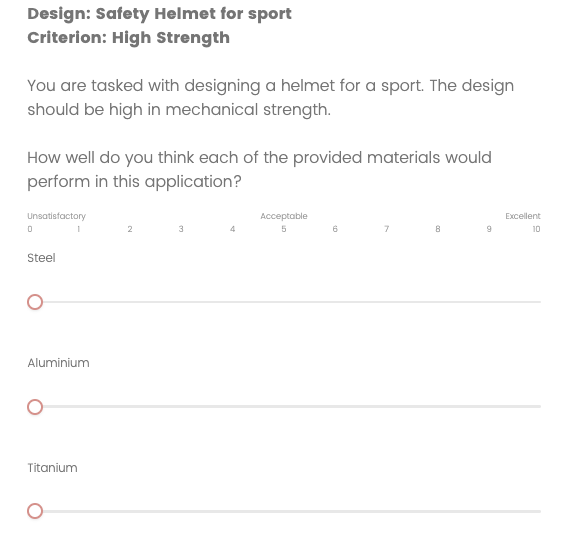}

\caption{\label{fig:survey-total}Example of a question asked on the survey, missing from the figure are the remaining sliders allowing participants to select a value from 1-10. The 16 questions asked in the survey were created by taking all possible combinations of design cases and criteria. In all the questions, the only change made is the highlighted text, Design and Criterion.}
\end{figure}

The survey queried participants across 4 design cases (\textbf{Kitchen Utensil Grip}, \textbf{Spacecraft Component}, \textbf{Underwater Component}, and \textbf{Safety Helmet for Sport}) and 4 design criteria (\textbf{Lightweight}, \textbf{Heat Resistant}, \textbf{Corrosion Resistant} and \textbf{High Strength}) combined in a full factorial experimental design to produce 16 scenario-based questions. In each question, participants were asked to score a set of materials (\textbf{steel}, \textbf{aluminum}, \textbf{titanium}, \textbf{glass}, \textbf{wood}, \textbf{thermoplastic}, \textbf{elastomer}, \textbf{thermoset}, and \textbf{composite}) on a scale from 0 to 10, with 0 being unsatisfactory in the specific application and 10 being an excellent choice. These material categories were chosen to cover a breadth of design use cases, to find a balance between high-level and low-level material categories, and to limit the length of the survey.  An example of a survey question is shown in Figure~\ref{fig:survey-total}. The survey also collected basic demographic information to ensure that participants had the necessary knowledge and background to provide strong preferences for material selection.

We utilized Qualtrics to design and deliver the survey. The survey was distributed to professionals who have worked as materials scientists, materials engineers, design engineers, or related fields, via the Autodesk Research Community. The survey remained accessible for 20 days.

\subsection{Dataset Organization and Description}
\label{sec:meth - data org}

The survey responses were extracted from Qualtrics and processed such that all identifiers of the participants and any information that can be linked to them are removed. The dataset is hosted in a HuggingFace repository and consists of three files: \texttt{AllResponses.csv}, \texttt{CleanResponses.csv}, and \texttt{KeyQuestions.csv}. 

KeyQuestions.csv is a key explaining what each column header means in the other two files. For example, column header \texttt{Q1\_Steel} corresponds to responses about steel to the question about designing a Kitchen Utensil Grip that should be Lightweight (i.e, Q1).

The survey was set up in such a way that if a respondent answered one or many question(s) but did not complete the survey, the response was still collected and the unanswered questions would remain blank in the final response even if the final survey was submitted or not. \texttt{AllResponses.csv} contains the responses of all respondents, regardless of whether the responses are complete or not. The \texttt{CleanResponses.csv} file is processed and constructed such that only responses are present for which a respondent completed all survey questions. Both files have the same number of columns but different number of responses because of the reason mentioned above. The number of responses (rows) in the file \texttt{AllResponses.csv} is 138 and the number of responses (rows) in the file \texttt{CleanResponses.csv} is 67.

The first column in both of these files is labeled \texttt{material\_familiarity}, and records that individual's level of familiarity with material selection. The next column is labeled \texttt{yrs\_experience} which is that individual's number of years of experience in that domain. The following 144 columns record the response to every combination design case and criteria for each material choice - resulting from 4 design cases, 4 design criteria, and 9 material choices.


\subsection{FAIR Principles}
\label{sec:meth - FAIR}
This data adheres to the FAIR principles of Findable, Accessible, Interoperable, and Reusable \cite{wilkinson2016fair}:

\begin{itemize}
    \item Findable: The dataset is hosted in a HuggingFace repository and is assigned a globally unique and persistent Digital Object Identifier (DOI) \cite{design_research_collective_2024}. The dataset card contains the details about the dataset and other metadata and the dataset is indexed in a searchable source HuggingFace\footnote{\href{https://huggingface.co/datasets/cmudrc/Material_Selection_Eval}{https://huggingface.co/datasets/cmudrc/Material_Selection_Eval}\label{HF}}. 
    \item Accessible: The data is accessible from the HuggingFace datasets library \cite{lhoest_datasets_2021} and is licensed with an MIT license \cite{noauthor_mit_2006} which is mentioned in the dataset card as well. The example usage of the dataset is shown in a Github repository\footnote{\href{https://github.com/cmudrc/MSEval}{https://github.com/cmudrc/MSEval}\label{Github Example}}.
    \item Interoperable: Data is provided in the ubiquitous \texttt{CSV} format to ensure dataset can be used across different platforms and tools and the metadata follows a standard schema. The data elements are also explained in section \ref{sec:meth - data org} and the example remapping is also provided in the Github repository.
    \item Reusable: The data collection method is mentioned in section \ref{sec:meth - online survey} and the versioning information can be found on HuggingFace. 
\end{itemize}


\section{Potential Applications} 
\label{sec:pa}
This dataset can be useful for engineering design research in several ways:
\begin{enumerate}
    \item Evaluating material selection algorithms and decision-making models: Researchers can use this dataset to develop and test various material selection algorithms, decision-making models, and knowledge-based systems to assist designers in the material selection process.
    \item Investigating the relationship between material properties and design requirements: The dataset provides information on a wide range of material properties, which can be used to study the correlations between material properties and specific design requirements, such as strength, stiffness, weight, or cost.
    \item Exploring the influence of material familiarity and experience on design decisions: The dataset includes information on the user's years of experience and material familiarity, which can be used to investigate how these factors impact material selection decisions and the overall design process.
\end{enumerate}

In general, this dataset can be a valuable resource for engineering design research, as it provides a comprehensive set of material-related information that can be used to advance the understanding and practice of material selection in conceptual mechanical design.

\section{Conclusion}
\label{sec:conc}
In this paper, we introduce MSEval, a novel benchmark specifically designed to assess the capabilities of machine learning and algorithmic models with respect to human-level cognition in the domain of material selection in conceptual design. This benchmark consists of survey responses from experienced professionals in the fields of material science and engineering, material selection, and design engineering. By focusing on scenario-based domain-specific tasks, MSEval enables a more meaningful evaluation of algorithmic model performance, bridging the gap between human cognition and machine capabilities in material selection.

This dataset is subject to several limitations. For instance, the data set covers only four design cases and four design criteria, which is not fully representative of the wide range of design scenarios encountered in real-world applications. As such, the dataset does not capture contextual factors that influence material selection, such as specific project constraints, economic conditions, regulatory requirements, or market trends. In addition, the nine material families used in this work are  high-level categories, and nuanced differences between sub-categories, such as different aluminum alloyss, are not addressed. The dataset also represents a static snapshot of opinions and practices at a particular point in time. Material science and engineering practices evolve, and the dataset may become outdated as new materials and technologies emerge. To address these limitations,  future iterations of the dataset should expand the range of design scenarios and criteria, increase the sample size and ensure a more diverse respondent pool, include additional contextual information about respondents and their decision-making processes, and use more sophisticated methods to capture material selection trade-offs.

We hope that our research drives innovation in developing more reliable AI assistants that can advance toward this niche domain. By identifying areas for improvement and better understanding the behavior of humans, our benchmark provides a foundation for creating AI systems that can handle complex human-centric tasks effectively.

\section*{Acknowledgments}
\label{sec:ack}
We would like to thank the anonymous survey participants in the Autodesk Research Community for their contributions to this research.


\nocite{*} 

\bibliographystyle{asmejour}   

\bibliography{asmejour-sample} 



\end{document}